\documentclass[twocolumn]{article}

\usepackage{arxiv}

\usepackage{lipsum}
\usepackage{graphicx}
\usepackage{float}
\usepackage[utf8]{inputenc} 
\usepackage[T1]{fontenc}    
\usepackage{hyperref}       
\usepackage{url}            
\usepackage{booktabs}       
\usepackage{amsfonts}       
\usepackage{nicefrac}       
\usepackage{microtype} 
\usepackage{amsmath}

\usepackage{dirtytalk}

\usepackage{amssymb}
\usepackage{times}
\usepackage{latexsym}

\usepackage{microtype}

\usepackage{dirtytalk}
\usepackage{amsmath}
\usepackage{amsfonts}
\usepackage{amssymb}
\usepackage{booktabs}
\usepackage{mathtools}
\usepackage{hyperref}
\usepackage[symbol]{footmisc}
\usepackage{hyperref}
\usepackage{comment}
\usepackage{stfloats}

\DeclareMathAlphabet{\mathpzc}{OT1}{pzc}{m}{it}
\newcommand{\ctx}{\boldsymbol C}    
\newcommand{\img}{\boldsymbol v}    
\newcommand{\ttl}{\boldsymbol x}   
\newcommand{\dsc}{\boldsymbol c}   
\newcommand{\prc}{p}   
\newcommand{\ctg}{k}   

\newcommand{\hst}{\boldsymbol h}  
\newcommand{\ut}{\boldsymbol u}          
\newcommand{\w}{\boldsymbol\omega}          

\newcommand{\pprc}{\Bar{\prc}}   

\newcommand{\ove}{\Phi}     
\newcommand{\balpha}{\boldsymbol\alpha} 
\newcommand{\emb}{\boldsymbol e}       
\newcommand{\Emb}{\Psi}       
\newcommand{\mem}{\boldsymbol m}       
\newcommand{\wgt}{w}       
\newcommand{\bwgt}{\boldsymbol\wgt}       
\newcommand{\rep}{\boldsymbol o}       

\newcommand{\hid}{\boldsymbol z}    
\newcommand{\stt}{\boldsymbol s}    
\newcommand{\act}{a}    
\newcommand{\pact}{\Bar{a}}    
\newcommand{\rat}{r}    

\newcommand{\app}{\boldsymbol\delta}    
\newcommand{\pap}{\boldsymbol\phi}    
\newcommand{\dcp}{\boldsymbol\theta}    

\newcommand{\Gin}{G}    

\title{Show, Price and Negotiate: A Negotiator with Online Value Look-Ahead}

\author{
        Amin~Parvaneh, 
        Ehsan~Abbasnejad, 
        Qi~Wu, 
        Javen~Qinfeng~Shi 
        and~Anton~van~den~Hengel\\
  Australian  Institute  for  Machine Learning, The University of Adelaide\\
  \texttt{\small{\{amin.parvaneh, ehsan.abbasnejad, qi.wu01, javen.shi, anton.vandenhengel\}@adelaide.edu.au}} \\
}

\begin{document}

\maketitle


\begin{abstract}
Negotiation, as an essential and complicated aspect of online shopping, is still challenging for an intelligent agent.
To that end, we propose the \emph{Price Negotiator}, a modular deep neural network that addresses the unsolved problems in recent studies by 
(1) considering images of the items as a crucial, though neglected, source of information in a negotiation, (2) heuristically finding the most similar items from an external online source to predict the potential value and an acceptable agreement price, (3) predicting a general price-based ``action" at each turn which is fed into the language generator to output the supporting natural language, and (4) adjusting the prices based on the predicted actions.
Empirically, we show that our model, that is trained in both supervised and reinforcement learning setting, significantly improves negotiation on the CraigslistBargain dataset, in terms of the agreement price, price consistency, and dialogue quality.
\end{abstract}


\section{Introduction}\label{sec:introduction}
Negotiation is an integral part of human interactions. 
It is a complex task that requires reasoning about the attitudes of the counterpart, mutual interests, and uttering convincing arguments and potentially appealing to sympathy. The prevalence of online shopping provides a test-bed for negotiation ability of artificial agents as human's advocate for the best deals. This artificial agent has to assess the photos of the advertised item, understand the textual content, estimate its true value compared to the others in the market, and conduct a dialogue with its counterpart to reach an agreement. 


Recently, Lewis et al.~\cite{Lewis_dealnodeal} pioneered negotiation as a specific form of dialogue systems in a DealOrNoDeal game where two artificial agents negotiate splitting of three items. Subsequently, He et al.~\cite{He2018_craigslist} used real human dialogues on Craigslist advertisements to learn a dialogue model of negotiations. In both cases, in par with other dialogue systems, various sequence-to-sequence (Seq-Seq) encoder-decoders are utilised to model negotiations.
Seq-Seq models (or more complex alternatives \cite{Devlin018_bert,Luong15_attention} for that matter) are effective tools for learning the correlation between words (e.g. co-occurrences) and potentially the goal. However, negotiation presents a unique set of challenges beyond word correlation that distinguishes it to that of the conventional dialogue systems. Subsequently, these methods struggle to attain some indispensable aspects of a negotiation including: (1) extracting and utilising information from multiple sources (e.g. photos, texts, and numerals), (2) predicting a suitable price for the products to reach the best possible agreement, (3) expressing the intention conditioned on the price in natural language, and (4) offering consistent prices.

\begin{figure}[t]
  \centering
  \vspace{33mm}
  \includegraphics[width=\linewidth]{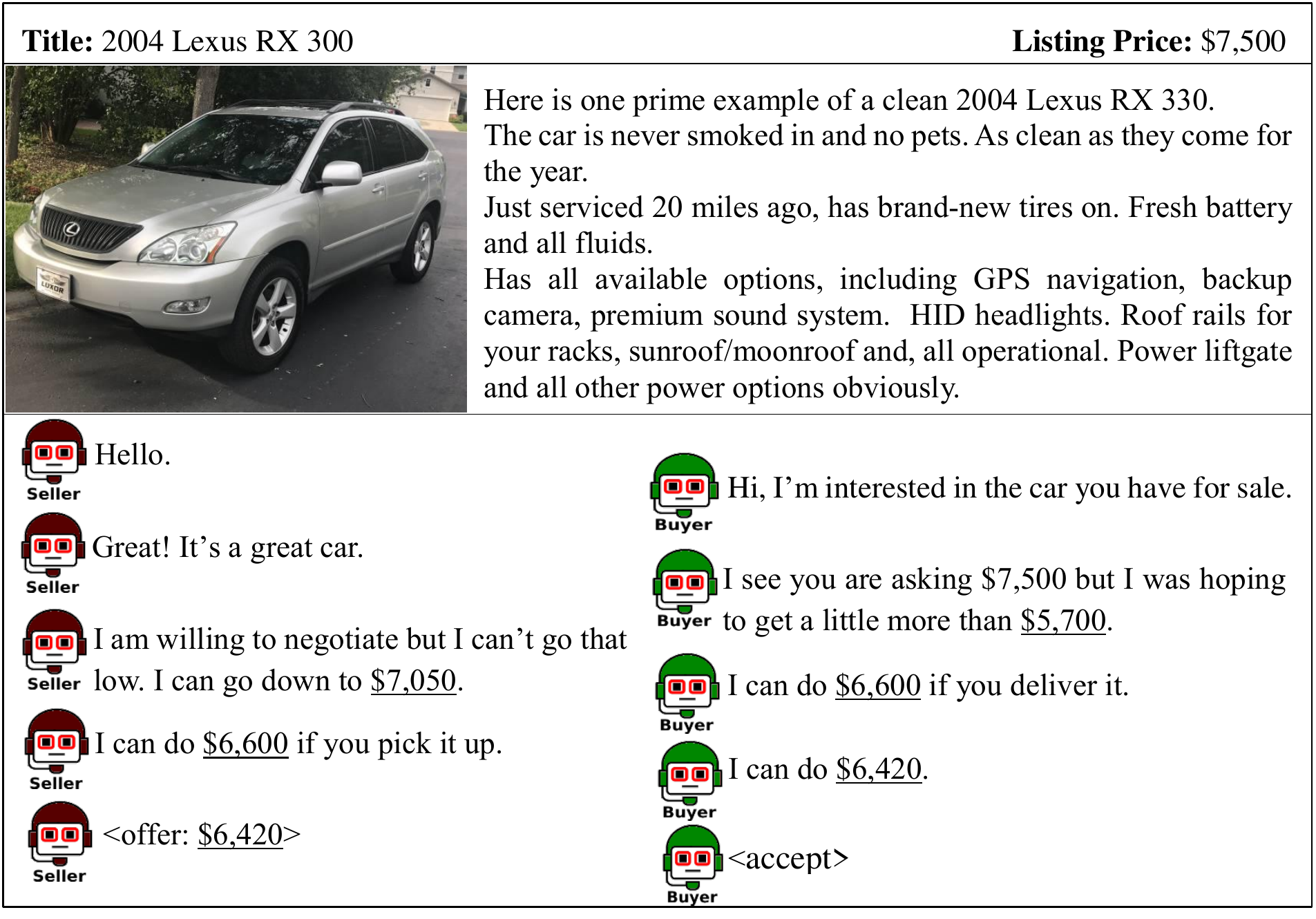}
  \vspace{-8mm}
  \caption{An example scenario of our \emph{Price Negotiator} from supervised and RL training. Two agents play the role of sellers and buyers in a visually-grounded bargaining game over an item. The agents need to uncover the underlying value of the item and the attributes of their counterpart (e.g. their assertiveness) to succeed.}
  \label{fig:first_sample}
  \vspace{-6mm}
\end{figure}

\begin{figure*} 
  \centering
  \includegraphics[width=\textwidth]{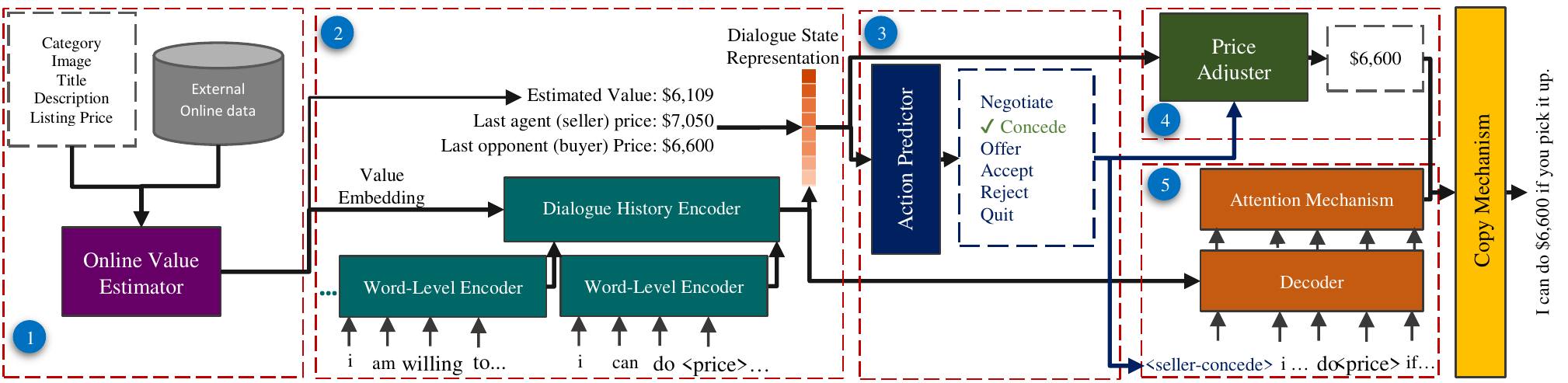}
  \vspace{-6mm}
  \caption{The diagram of our \emph{price negotiator} that consists of five main modules: (1) online value estimator (OVE), (2) hierarchical recurrent negotiation encoder (HRNE), (3) action predictor, (4) price adjuster, and (5) language decoder.}
  \label{fig:VisualNegotiator}
  \vspace{-5mm}
\end{figure*}

In this paper, we propose a \emph{price negotiator} to address the aforementioned problems. Our negotiator, inspired by the modular needs of a negotiating agent, comprised of five main units particularly tailored for shopping: (1) \emph{online value estimator (OVE)}, (2) \emph{hierarchical recurrent negotiation encoder (HRNE)}, (3)  \emph{action predictor} controller, (4) \emph{price adjuster} and (5) \emph{language decoder} 
(see Figure~\ref{fig:VisualNegotiator} for details). For \emph{OVE}, motivated by human behaviour, before starting negotiation we find similar items in online stores--simulating market evaluation. This is done by learning an embedding for the \emph{textual} (title and description) and \emph{visual} content of the listings and using a matching network to choose the most similar ones to the current item in the negotiation. Hence, an estimate of how much the item valued is prognosticated that allows the agent to uncover how demanding an item is and whether it's worth the listing price.

Subsequently, in \emph{HRNE} the counterpart's dialogue is encoded conditioned on the content of the advertisement and the agent's belief of its value. This is a significant and distinguishing aspect of our approach since OVE and HRNE effectively disentangle the value of an item from the language model. The output of this step is a dialogue state representation (encoding a combination of dialogue history representation, last prices proposed by the agents, textual and visual inputs and the estimated value) from which \emph{action predictor} decides on the next step for the negotiation. In a nutshell, action predictor decides on continuing with the intention of convincing the counterpart, conceding, offering a price, accepting their terms or quitting. If the decision is to change the offer, then our \emph{price adjuster} proposes a new price. From the state representation and the predicted action, our \emph{language decoder} generates the appropriate language to convey the intentions of the agent. In any case, we use \emph{copy mechanism} \cite{luong-etal-2015-addressing_rare_word} to combine the new offered price to that of the appropriate negotiating words to utter. 

We evaluate our proposed model on CraigslistBargain~\cite{He2018_craigslist} which provides human-generated negotiations in various scenarios using Craigslist advertisements. Our experiments show that not only the language quality of the generated utterances from our approach outperforms the baselines, 
the prices are consistent and the agreed price is more similar to that of humans. Moreover, we show \emph{reinforcement learning} \cite{Williams_reinforce}--that has become increasingly popular with dialogue systems--also improves our model's performance. We also run several human studies to evaluate our negotiator.


In summary, our main contributions are as follows:
\begin{enumerate}
\item We propose a novel AI agent that performs negotiation for the best price for either a seller or a buyer. It utilises both visual and textual content for decision making, follows a consistent and human-like pricing strategy and, as our experiments show, outperforms the baselines on both language quality and agreement price.

\item Our negotiator, unlike its counterparts, is able to find the relevant online items to accurately predict its potential agreement price.
This enables scalable and commercially viable applications and reduces human bias and inconsistency.  
\end{enumerate}

\section{Related Work}

\subsection{Goal-Oriented Dialogue}
Goal-oriented dialogue systems have a long history in natural language processing (NLP). Recently, researchers suggested to define a goal in open-domain dialogues to improve the consistency and engagement of the agent \cite{tang-etal-2019-target}. Additionally, multi-modal dialogue systems have gained strong interests in speech recognition \cite{spokendial2_tmm} and computer vision communities\cite{Liao2018_multimodal, multimodal_aaai_2018}. Specifically, visual goal-oriented dialogue systems have got the popularity by introducing miscellaneous tasks including \say{GuessWhat?!}\cite{Vries2017GuessWhatVO,lee2018largescale}, \say{Image Guessing} \cite{Das2017LearningCV}, \say{MNIST Counting Dialogue} \cite{Lee2018AnswererIQ}, \say{Visual Dialogue} \cite{Das2017VisualDialog} and \say{CLEVR-Dialogue} \cite{Kottur2019CLEVRDialog}. However, since the machine can play just one role (either questioner or answerer) in most applications, they are Visual Question Answering problems by nature rather than two-way, interactive dialogue systems \cite{Wu2017VisualQA, Das2017VisualDialog, Kottur2019CLEVRDialog}. In this paper, we focus on \emph{Visual Negotiation} where the model is evaluated interactively in negotiations either with humans or with another model.

 Generally, dialogue systems can be categorised into \emph{collaborative} and \emph{competitive} systems. In a collaborative dialogue environment, agents can help each other to reach a common goal. Applications include trip and accommodation reservation \cite{Layla2017_Frames,Wei18_airdial}, information seeking \cite{Siva2018_coqa,whatknow,spokendial1_tmm}, mutual friend searching \cite{He2017_collaboraive}, navigation \cite{Ond2016,Vries2018_TalkTW}, fashion product recommendation \cite{Liao2018_multimodal}, disease diagnosis \cite{Xu2019EndtoEndKR}, addressee detection \cite{adressee_tmm}, emotion detection \cite{Majumder2018DialogueRNNAA}, and even donation persuasion \cite{wang-etal-2019-persuasion}.
In contrast, in a competitive dialogue environment, agents must negotiate to achieve an agreement based on their individual goals. Their goals are often opposite to each other. \say{Settlers of Catan} \cite{Cuayhuitl2015StrategicDM} and DealOrNoDeal \cite{Lewis_dealnodeal} are two frontier tasks defined as competitive dialogues. Very recently, a new negotiation dataset is introduced by crawling tangible negotiation scenarios from the Craigslist website and collecting seller-buyer dialogues for each scenario \cite{He2018_craigslist}. 
Although our work is built on top of the same dataset, there are significant differences: (1) \textit{we propose to use the photos of the item as an important source of knowledge which was neglected in \cite{He2018_craigslist}}; (2) \textit{before the negotiation, we prognosticate an ideal agreement price by analysing other similar items on online stores}; (3) \textit{we aim to estimate and refine the price in a consistent manner, and produce human-like dialogues. }


\subsection{Dialogue Systems Design}
Goal-oriented dialogue systems can be designed in a component-based fashion or end-to-end. In a component-based fashion, it typically has three separate modules: (1) natural language understanding (NLU) unit that maps an utterance into semantic slots to be understood and processed by the machine, (2) dialogue manager (DM) which selects the best action according to the output of NLU, and (3) natural language generator (NLG) which produces a meaningful response based on the action chosen by DM, either by looking at a set of possible responses for that action or by using a statistical machine learning language model \cite{Hongshen_dial_survey, Victor_conv_infer}. 

To overcome the complexity and bypass the reliance on human-crafted information retrieval rules in component-based approaches, end-to-end systems have been proposed in recent years \cite{Tsung_net_dial, Antoine_end_to_end_dial, Bhuwan17_end2end, Xiujun17, Jiwei2017_adversarial, Sordoni2015_hred, Xing2018_hran, Ond2016}. These systems often use an Seq-Seq architecture consisting of an encoder which receives the previous utterance(s) and encode them into a latent representation based on which the decoder can predict and generate the next utterance. 
In the end-to-end model proposed by He et al. \cite{He2018_craigslist} for the negotiation, prices are embedded similar to other words in the utterance. Since the range of the prices are broad and there is not any pre-trained embedding for them, their embedding is learned through the model training. In addition, the generated prices are inconsistent since they were produced based on the correlation with other words rather than the true underlying value of the item. Furthermore, this way of embedding the prices adds more complexity to the model and leads to weaker language model. 
In this research, we show that eliminating the prices from the dictionary of the model, can help the language model to generate better dialogues. We propose an end-to-end modular approach in which we predict the price and the supportive language separately from different heads of the network. 

\section{Price Negotiator}

\subsection{Problem Definition}
The problem we consider is that of having two agents, namely a seller and a buyer, negotiating on the price of an item which is identified by an image, textual title and description. The items are classified into various categories as is the common practice in the online shopping websites.
The seller advertises an item with a \emph{listing} price and most likely agrees to offers closest to this value. The buyer on the other hand has a \emph{target} price which is lower than the seller's listing. While the buyers know the listing price, their target price is not revealed to the seller. It should be noted that a negotiation may end without an agreement.


Each negotiation scenario consists of an advertised item $i$ by providing its context information ${\ctx}_i=\{{\img}_i, {\ctg}_i, {\ttl}_i, {\dsc}_i, {\prc}_{0,i}\}$, where ${\img}_i$ represents its visual cue/feature (i.e. photo), ${\ctg}_i$ is the category in which the item has been advertised, ${\ttl}_i$ is the title of the advertisement, ${\dsc}_i$ is the description provided for the item, and ${\prc}_{0,i}$ is the listing price suggested by the seller. Additionally, at each dialogue turn $t$, a sequence of utterances in previous turns is available as the dialogue history ${\hst}_{i,t}=\{{\ut}_{i,0}, {\ut}_{i,1},\ldots, {\ut}_{i,t-1}\}$. It is noticeable that each utterance is a sequence of words (tokens) ${\ut}_{i,t}=\{\w_{i,t,0}, \w_{i,t,1},\ldots, \w_{i,t,L-1}\}$, where $L$ represents the maximum length of each utterance, and each word is represented as a $d$-dimensional vector.

At $t$-th round of negotiation, the agent generates the $j$-th token conditioned on the context information ${\ctx}_i$, the dialogue history ${\hst}_{i,t}$, and the previously generated tokens $\{\w_{i,t,0}, \w_{i,t,1},\ldots, \w_{i,t,j-1}\}$. 
The objective is to as closely as possible mimic the behaviour of a human in negotiation. Consequently, the prices agreed upon by an agent has to be as similar as possible to that of the human using convincing arguments. 

\begin{figure}[t]
  \centering
  \includegraphics[width=\linewidth]{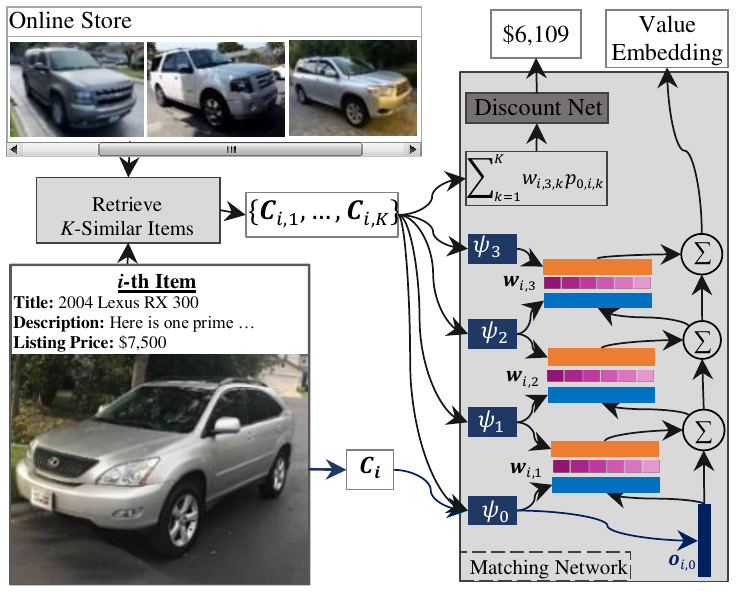}
  \vspace{-6mm}
  \caption{The architecture of \emph{Online Value Estimator} that finds $K$-similar items in an online store. Comparing their multi-modal contextual features $\{{\ctx}_{i,1},\ldots,{\ctx}_{i,K}\}$ with those of the given item ${\ctx}_{i}$, it prognosticates a fair agreement price to be considered in the negotiation. See Figure~\ref{fig:MultimodalEmbedding} for more details about multi-modal embeddings ${\Emb}_0, \ldots, {\Emb}_3$.} 
  \label{fig:OVE}
  \vspace{-5mm}
\end{figure}


\subsection{Online Value Estimator}

One of the essential skills in negotiation is to have a good estimation of the real value of the item. Humans usually search through different shopping websites to find similar items and compare their attributes and listing prices with those of the given item. Motivated by this, we designed the \emph{online value estimator} (OVE), a deep neural network that can make a precise value prediction (Figure~\ref{fig:OVE}).

Given the context information ${\ctx}_i$ of the item $i$, the OVE component predicts a scalar value for the agreement price. This estimation is based on both visual features of the item, extracted from its photo ${\img}_i$; its textual features extracted from its category ${\ctg}_i$, title ${\ttl}_i$ and description ${\dsc}_i$; and its listing price ${\prc}_{0,i}$. 
Generally, the OVE component aims at minimising the difference between the predicted price ${\pprc}_i$ and the ground-truth real agreed price ${\prc}_i$. The ground-truth real price is calculated as the average of all agreed prices in human-human negotiations over the given item in the dataset. To predict a price we learn a deep neural network $\ove$ parameterised by $\balpha$ through the minimisation of the following loss:
\begin{equation}
\small{
    \boldsymbol{\ell}_{ove}=\sum_{i=1}^{N}|\ove({\img}_i, {\ctg}_i, {\ttl}_i, {\dsc}_i, {\prc}_{0,i};\balpha)-{\prc}_i|,\label{eq:ipe}}
\end{equation}
where $N$ represents the number of items in the training set.

The price is predicted in a three-stage process. First, the extracted features of the items are used to find the $K$-similar items from an online source of advertised items. The similarity between two items is defined as a combination of cosine similarities between their visual and textual features and the normalised abstract similarity between their listing prices. 

\begin{figure}[t]
  \centering
  \includegraphics[width=\linewidth]{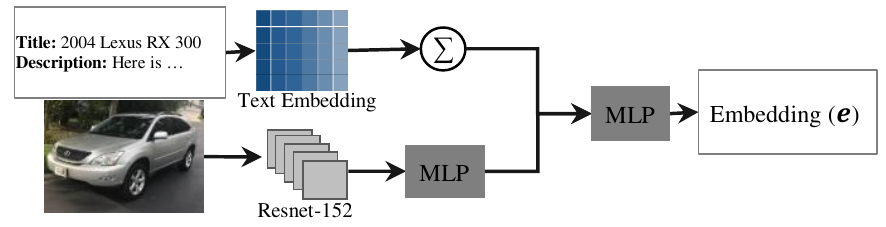}
  \vspace{-5mm}
  \caption{Multimodal Embedding. The sum of embeddings of the textual resources (title and description)  is concatenated with the down-sized visual features obtained from a pre-trained network (ResNet-152~\cite{Resnet_2016}). The concatenated vector is passed through a 2-layer MLP with ReLU activation to create the multimodal embedding.}
  \label{fig:MultimodalEmbedding}
  \vspace{-5mm}
\end{figure}

Second, the \emph{matching network}, a deep neural network with a structure akin to memory networks \cite{Sukhbaatar2015_MemNet}, takes these items and measures their importance in valuing the item. 
It worth mentioning that in contrast to \cite{Sukhbaatar2015_MemNet}, where they only embed text inputs, we propose \emph{multimodal embedding} (Figure \ref{fig:MultimodalEmbedding}) that embeds the visual and textual features of the given item $i$ into a $d$-dimensional representation ${\emb}_{i}$.  
Specifically, our proposed matching network consists of 3 attention layers and 4 multimodal embeddings. At each layer $l$ the the correlation between the previous representation of the given item ${\rep}_{i,l-1}$ and multimodal embeddings of related $K$ similar items ${\mem}_{i,l}=[{\emb}_{i,l,1},\ldots,{\emb}_{i,l,K}]$, which are extracted from $(l-1)$-th multimodal embedding ${\Emb}_{l-1}$, is calculated as follows: 
\begin{equation}
\small{
    {\bwgt}_{i,l}=\text{Softmax}({\rep}^{\intercal}_{i,l-1} {\mem}_{i,l})}.
\end{equation}
Afterwards, the output of the layer (the item representation ${\rep}_{i,l}$), is calculated based on the following equation:
\begin{equation}
\small{
    {\rep}_{i,l}={\rep}_{i,l-1} + \sum_{k=1}^K ({\wgt}_{i,l,k} {\emb}_{i,l,k})}.
\end{equation}
Please note that the initial item representation ${\rep}_{i,0}$ also comes from the first multimodal embedding.

Finally, the correlation weights from the last layer of the network are multiplied by the listing prices of the corresponding similar items to achieve an estimated value ${\pprc}_i$ for the given item. It worth to mention that since this value is calculated from the listing prices and the target is the agreement price, we pass the output through another fully connected layer, which we name it \emph{discount net}, to estimate the final value for the item.

\subsection{Hierarchical Recurrent Negotiation Encoder}


One of the problems in conventional negotiation models is that they include price values (real numbers) in the vocabulary and treat them like ordinary words in the dialogue. This deters the intelligent agent from understanding the numerical meaning of the prices, and entangles the strategies for generating words and prices together. As a result, the prices generated in the dialogue, especially at final offering turn, are inconsistent in most cases. 

In our price negotiator we devise a novel \emph{hierarchical recurrent encoder} in which the prices in the utterances are replaced with a fix token (\texttt{<price>}) to be later replaced with the generated ones. In a hierarchical structure~\cite{Sordoni2015_hred,Xing2018_hran,Ond2016}, our model encodes utterances in two levels: a \emph{word-level encoder} that is an RNN network 
($f^{we}:\mathbb{R}^{L\times d}\rightarrow \mathbb{R}^{d}$) mapping the word embedding of $t$-th utterance (a sequence of maximum $L$ words) into a $d$-dimensional vector (${\hid}_{i,t}^{we}$) as the word-level representation of the utterance; and a \emph{dialogue history encoder} that is another RNN network 
($f^{he}:\mathbb{R}^{(t-1)\times d}\rightarrow \mathbb{R}^{d}$) which at each turn $t$ receives word-level representation of the previous utterances as the input and maps them to a $d$-dimensional vector (${\hid}_{i,t}^{he}$). Since this representation should be conditioned on the value estimation resulted from OVE, we feed the output of the last layer of the matching network into this RNN as the initial hidden state.

Apart from the dialogue history representation, the last prices suggested by the agent ${\prc}_{i,t-1}^a$ and the opponent ${\prc}_{i,t-1}^o$ and the estimated price ${\pprc}_i$ are embedded into a vector which represents the dialogue state ${\stt}_{i,t}$ (more details in section \ref{details_emb}). This vector will then be used by other components to decide about action and prices that should be considered. 

\subsection{Action Predictor}

The action predictor module is a multi-layer perceptron (MLP) that predicts the next action ${\act}_{i,t}$ should be taken by the agent according to the dialogue state ${\stt}_{i,t}$ at round $t$ of the negotiation. In contrast to \cite{He2018_craigslist} who tried to predict coarse intents based on intent encoding, we suggested to predict extremely simpler actions, which are based on the price. Actions defined in our framework are:
\begin{itemize}
    \item \emph{Negotiate} tells the agent that it should continue the negotiation without changing the price.
    \item \emph{Concede} determines that the agent should make a concession on its previously proposed price. In other words, the buyer should increase its suggested price and the seller needs to decrease its asking price when this action is predicted.
    \item \emph{Offer} suggests that the agent should propose a final offer and wait for the response from its counterpart.
    \item \emph{Accept} means that the agent should accept the official offer suggested by the opponent and terminate the negotiation successfully.
    \item \emph{Reject} clarifies that the agent should reject the proposed offer.
    \item \emph{Quit} means the agent should abandon the negotiation.
\end{itemize}

In the supervised training setting, this neural network learns parameters $\app$ that better imitate human-like actions by minimising this loss function:
\begin{equation}
    \small
        {\boldsymbol{\ell}_{ap}= \sum_{i=1}^N\sum_{t=1}^{T_i} {-\log {{p}({\act}_{i,t}|{\stt}_{i,t};\app)}},}
\end{equation}
where $N$ and $T_i$ represent the number of training dialogues and the number of agent's turns in each dialogue respectively. 

\subsection{Price Adjuster}

Proposing a reasonable price at each stage of the dialogue is fundamental for a negotiation agent. Our price adjuster module can make consistent price suggestions that lead the agent to reach the best possible agreement. This module is invoked only if the action predictor decides to concede or make an offer. In either case, the price adjuster, an MLP with parameters $\pap$, predicts the ratio ${\rat}_{i,t}$ from which the agent should concede.  This prediction is based on the current state of the dialogue ${\stt}_{i,t}$ and the action predicted by the action predictor ${\pact}_{i, t}$. We discretise the price change ratio into six categories (more details in section \ref{details_price_adjustment}) and optimise the network using this loss function:
\begin{equation}
\small{
    \boldsymbol{\ell}_{pa}=\sum_{i=1}^N\sum_{t=1}^{T_i} {-\log {{p}({\rat}_{i,t}|{\stt}_{i,t},{\pact}_{i, t};\pap)}},}
\end{equation}

\subsection{Language Decoder} 
Language decoder is an RNN that generates a sequence of words as the next utterance based on current dialogue state ${\stt}_{i,t}$ and the predicted action ${\pact}_{i, t}$. To that end, we initialise its hidden state with the last hidden state from the dialogue history encoder. Additionally, we condition the starting token on the selected action, by defining different tokens for different actions. We then train an agent able to play both seller and buyer roles by defining different start tokens for the buyer and the seller.

In order to encourage the output to pay more attention to the  most important parts of various available information sources, a global attention mechanism \cite{Luong15_attention,gu-etal-2016-copy} is applied to the outputs of the language decoder. This helps the system to ask or answer questions for different sources including the title, description and the outputs of word-level encoder for previous utterance. 

To map the outputs of the model to a probability vector of our vocabulary size, a linear function (generative layer) and a $\text{LogSoftmax}$ is applied to the output of the model. With language decoder we find the parameters $\dcp$ of the RNN to maximise the likelihood of each word,
    \begin{equation}
    \small{
        \boldsymbol{\ell}_{ld}=\sum_{i=1}^N\sum_{t=1}^{T_i}\sum_{j=1}^L {-\log {{p}(\w_{i,t,j}|{\ctx}_i, {\hst}_t, \w_{i,t,0},\ldots, \w_{i,t,j-1};\dcp)}}.}
    \end{equation}

\subsection{Copy Mechanism} 
We disentangle prices from other words during the encoding and decoding by replacing prices in utterances with a fix token (\texttt{<price>}). While we encode the current proposed prices separately, the decoder only predicts the price location in the generated utterance. Similar to \emph{copy mechanism} utilised in machine translation and question answering \cite{luong-etal-2015-addressing_rare_word,gu-etal-2016-copy}, we replace the price point predicted by the language decoder module with the value calculated by the price predictor module to create the final output.

\subsection{Overall Objective}
The final objective function for the price negotiator model is to minimise the combination of losses introduced for each component.
\begin{equation}
\small{
        \boldsymbol{\ell}=\boldsymbol{\ell}_{ove}+\boldsymbol{\ell}_{ap}+\boldsymbol{\ell}_{pa}+\boldsymbol{\ell}_{ld}.}
\end{equation}


\subsection{Reinforcement Learning}
We use reinforcement learning to encourage our \emph{Price Negotiator} agent to improve by employing self-play (i.e. two instances of our model play buyer and seller roles and negotiate with each other). 
Specifically, once supervised training of the network is done, the action predictor and the price adjuster are fine-tuned using REINFORCE algorithm.
We assign a role to an agent (say seller) and let it negotiate against another (e.g. buyer) for a given scenario (i.e. image, title, description, listing price). At the end of negotiation, we evaluate the performance by providing a \emph{reward} signal. Our reward signal 
measures how successful the agent was according to the distance between the agreed price and the estimated price ${\pprc}_i$ predicted by the OVE. The motivation for the reward signal is to intrigue the agent to mimic human's strategy and achieve the same agreement price.
Thus, the action predictor network is updated by back propagating the following signal: 
 \begin{eqnarray}
     \sum_{i=1}^N\sum_{t=1}^{T_i}{\log {{p}({\pact}_{i,t}|{\stt}_{i,t};\app)}} {\Gin}_i,
 \end{eqnarray}
where ${\Gin}_i$ represents the total reward for negotiation $i$. The same update is applied on price adjuster component, which is eliminated for the brevity.  

\begin{figure*}[t]
  \centering
  \includegraphics[width=\textwidth]{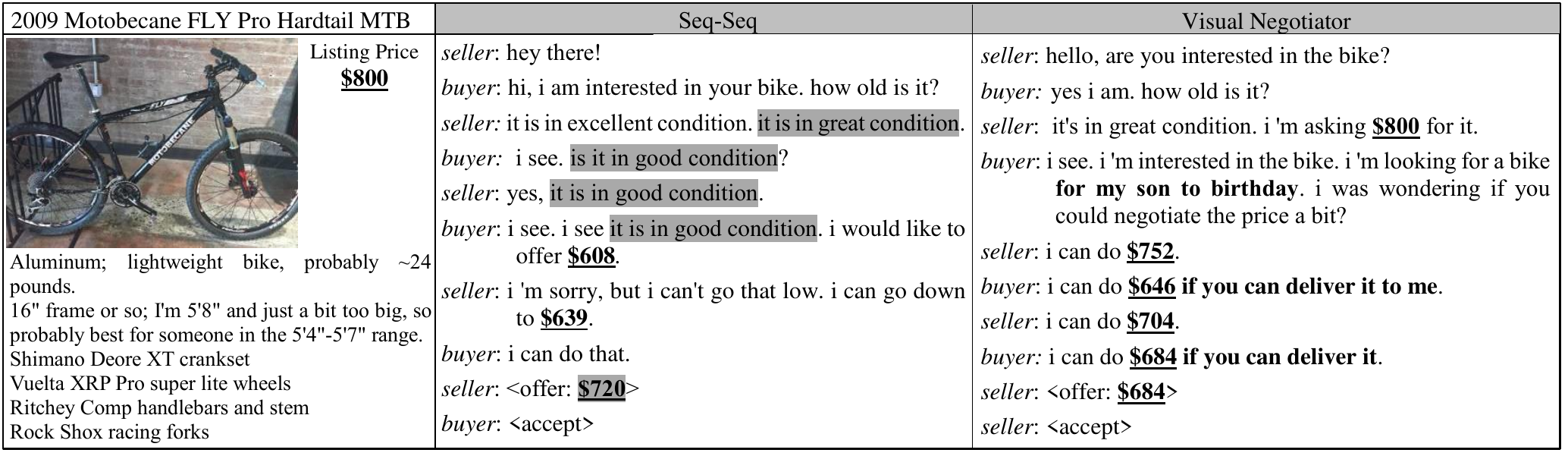}
  \vspace{-6mm}
  \caption{\small{Examples from two models. The left-hand side dialogue is generated from a simple Seq-Seq model which makes a mistake in its offer (from $\$639$ that was agreed upon moves to $\$720$) and produces repetitive, dull responses. On the other hand, our \emph{Price Negotiator} model, the right-hand side dialogue, creates a linguistically more diverse and price-wise reasonable dialogue and uses different negotiation strategies to persuade the opponent.}}
  \label{fig:second_sample}
  \vspace{-5mm}
\end{figure*}

\section{Implementation Details}

\subsection{Dataset}
All the experiments are performed on the CraigslistBargain dataset \cite{He2018_craigslist}. 
It contains 4,219 training dialogues, 471 evaluation dialogues and 500 test dialogues which are created based on $891$, $103$, and $134$ different items respectively. 

To make the training process of OVE simple and fast, we simulated an online external data source. To that end, we scraped $10,586$ items advertised on Craigslist website and made a local source accessible by our online value estimator. It is notable that we set the number of selected items from the online source $k$ to 32 in all experiments. 



\subsection{Embeddings} \label{details_emb}

In all the experiments, we use 300-dimensional vectors as the embedding for each word from pre-trained GloVe embedding~\cite{pennington2014glove}. In order to extract the features from the images, we utilised Resnet-152~\cite{Resnet_2016} pre-trained on ImageNet dataset, which has shown exceptional performance in various object detection problems. We simply replaced its last fully-connected layer with another one to produce a $300$-dimensional vector representing the image features. 

The hierarchical recurrent negotiation encoder maps prices (either the agent price, the opponent price, or OVE estimated price) into a 7-dimensional one-hot vector which will be concatenated with the last hidden state of its dialogue history encoder to represent the dialogue state. To that end, similar to \cite{He2018_craigslist}, prices are first normalised separately for each agent so that 1 is the agent's target price and 0 is their bottom-line price. Defined in the negotiation scenario, the bottom-line price for the seller is the lowest price he/she is supposed to sell the item while for the buyer it is the highest value they should pay for the item. 
It worth mentioning that these bottom-line prices are not strict and agents can propose and agree on values outside of this range. 
After the normalisation, price range between 0 and 1 is segmented equally to 5 parts representing 5 classes and other two classes belong to values lower than 0 and higher than 1. Please note that prices lower than the seller's bottom-line price or higher than buyer's bottom-line price are represented as negative values for them.


\subsection{Training Settings}
All RNNs used as encoder or decoder are 2-layer LSTMs with $300$-dimensional hidden states. 
Action predictor and price decoder networks have the same network architecture, a 4-layer fully-connected network with ReLU activation functions. We also applied a dropout with a rate of 0.3 to all parts of our architecture. 

Parameters of the models are optimised using Adam with the learning rate set to 1e-3 in first 20 epochs and then decayed to 1e-4 for another 320 epochs. The batch size is set to 128 in all experiments. 

It worth to mention that since the number samples for each class in both action prediction and price decoding tasks are imbalanced, we used weighted Cross-Entropy loss function where $\frac{1}{\sqrt{frequency(C_i)}}$ are calculated for each class $C_i$ and are used as the weights after normalisation.

Except for hierarchical dialogue encoder and the language decoder which are trained together, we trained other modules separately during the supervised learning process. Afterwards, for RL, we only optimised the action predictor and price decoder parameters for 5000 episodes using a learning rate of 1e-4.

\subsection{Price Adjustment}  \label{details_price_adjustment}
In the experiments, the price adjuster branch of the negotiation model makes a multi-class decision about the ratio with which the agent should retreat. Specifically, the agent always begins from the listing price if it acts as the seller and $50\%$, $70\%$, or $90\%$ of the listing price (depending on the scenario) if it acts as the buyer. Then, at each turn, the price decoder decides how much the current price should be changed. Price changes are categorised into 6 classes representing the change ratio which is $0\%$, $20\%$, $40\%$, $60\%$, $80\%$, or $100\%$ of either the listing price minus the target for the buyer or listing minus the $70\%$ of list for the seller. If the price decoder decides on altering the price, and the generated utterance contains a price token, it decreases (if being a seller) or increases (if being a buyer) the current price by the predicted ratio.

\subsection{Evaluation Metrics}


In this task, a negotiation is successful if the agents reach an agreement (by accepting a final offer from the opponent). However, the successful rate is not a suitable evaluation metric in this case because the `deal' might be a bad one even it was accepted. Instead, 
we defined various dialogue evaluation metrics which can be categorised into three groups: (1) metrics that evaluate the language quality (human-likeness) of the generated dialogue, (2) metrics that evaluate the pricing strategy of the model, such as the difference between the machine agreed price and the ground truth price, and (3) human studies.

\textit{Language Metrics.} 
In addition to BLEU score that measures how similar are the sequence of words generated in a machine-machine negotiation with those from human dialogues, we introduce Intent-BLEU (IBLEU), a new metric to measure the similarity of the intents taken by a machine with those taken by humans. 
More specifically, we extract the intents of each dialogue turn in a machine generated dialogue using the information retrieval approach introduced by \cite{He2018_craigslist}. For instance, the intent of saying \say{hello.} is \texttt{intro} and the intent of saying \say{I can’t go that low. I can go down to \$7,050.} is \texttt{counter-price}. Then, the extracted intents are concatenated to create a sequence of intents that will be compared with those extracted from human generated dialogues. This is basically done by calculating the clipped $n$-gram precision (for a maximum order of 4) of the generated sequence. The higher values of this metric shows the higher level of similarity with human.



Apart from IBLEU, we applied various word-level and sentence-level metrics to show the richness of the generated dialogues. We calculate the number of distinct sentences produced by the model and scale them by the total number of sentences as another new metric to show the language quality of the dialogue model. We also calculate the same metric at word-level to show the diversity of the lexical used by the model. And shorter dialogue length normally means the agent can make the deal in a shorter time.


\textit{Pricing Metrics.}
Two important metrics that measure the mistake ratio of the pricing are price inconsistency and offer inconsistency. When the seller proposes a price that is higher than the price previously suggested by themselves or is lower than the price offered by the seller, we consider it as an inconsistent pricing (see Figure~\ref{fig:second_sample}). 
We also calculate the average distance between the agreed prices from the human's agreed prices as another important measure to evaluate the pricing strategy of the model.




\textit{Human Studies.}
Apart from automatic evaluation, we measured the performance of our price negotiator model using human evaluations. We designed three  experiments to evaluate the human-likeness, language richness, and pricing quality of the negotiation models based on both third-party and interactive human evaluation.

\vspace{-2mm}
\begin{itemize}
    \item \textit{Turing test:} In this study, given a random dialogue generated from a negotiation model, the participant is asked to clarify whether or not the dialogue is generated by humans.
    \vspace{-2mm}
    \item \textit{Comparative test:} During this test, two dialogues generated based on the same scenario, one from our price negotiator and another from the baseline model, are shown to the contributor. The participants are tasked with choosing the best negotiation according to the language and pricing qualities.
    \vspace{-2mm}
    \item \textit{Interactive test:} Similar to~\cite{He2018_craigslist}, we put our price negotiator model along with the baseline model online and asked human volunteers to have a negotiation with a randomly chosen agent. At the end, they are asked to assess the quality of the agent in terms of human-likeness, language fluency and coherency, and pricing competency. 
\end{itemize}
\vspace{-4mm}

In our experiments, we gathered 400 Turing test, 400 comparative, and 20 interactive chat evaluations from 20 participants. It is  worth noting that the scenarios and generated chats are randomly selected from the test set.

\begin{table}[t]
 \caption{Value estimation results. Values in the table demonstrate the average divergence of the value estimations from the humans' agreed prices in the test set.}\label{tab:VE_results}
\vspace{-2mm}
  \centering
  \resizebox{0.85\linewidth}{!}{
  \begin{tabular}{p{1.2cm}p{1.2cm}p{1.2cm}p{1.2cm}|p{1.2cm}}
    \toprule
    \textbf{Category}&\textbf{Averaging}&\textbf{AVE}&\textbf{O-KNN}&\textbf{OVE}\\
    \midrule
    Bike&           \$475 &     \$422 &    \$55 &      \textbf{\$27}\\
    Car&            \$3,452 &   \$3,887 &   \$547 &     \textbf{\$495}\\
    Electronics&    \$114 &	    \$69 &	    \$14 &	    \textbf{\$6}\\
    Furniture&      \$191 &	    \$167 &	    \$26 &	    \textbf{\$20}\\
    Housing&        \$433 &	    \$458 &	    \$205 &	    \textbf{\$129}\\
    Phone&          \$112 &	    \$125 &	    \$21 &	    \textbf{\$20}\\
    \midrule
    Overall&        \$993 &	    \$898 &	    \$155 &	    \textbf{\$123}\\
  \bottomrule
\end{tabular}
}
 \vspace{-5mm}
\end{table}

\section{Results}

\subsection{Value Estimation Experiments}

To have a better understanding about the value estimation ability of our model, we calculate the average normalised divergence of the value estimations from humans’ agreed prices in the test set, for each product category. Table ~\ref{tab:VE_results} shows the results and several baseline models that are implemented for comparison. The simplest method is to use the average value of each category as the estimated value. We call this approach \emph{averaging}. We also train an attention neural network named \emph{attention value estimator} (AVE) that takes the visual and textual features of the item and outputs the predicted value. The third baseline is the average price of the $k$-similar items found from external online source and we name it online k-nearest neighbours (O-KNN). Since O-KNN approach is using listing prices and to have a fair comparison with our model which enjoys a discount network, we applied the average discount ratio from the discount network (around $11\%$) over the estimated values of this approach. 

From the Table ~\ref{tab:VE_results}, we can see the model proposed for online value estimation (OVE) can prognosticate an accurate agreement price for an item and beats all other baselines significantly. Using external sources and comparing item features with other items is useful, as the divergence of the predicted prices with the real agreement ones drop significantly in O-KNN approach. More importantly, the divergences drop extremely when similar items are matched with the given item using OVE model. 


\subsection{Negotiation Dialogue Evaluation}

In order to compare our negotiator with other baseline models, we train three state-of-the-art methods that treat the prices as words. 
The first two models are trained to match the method proposed in \cite{He2018_craigslist} on CraigslistBargain. The first one is a simple sequence-to-sequence model, SL(word), and the second one is a modular approach (SL(act)+rule) which has applied various human-crafted rules to repeat utterances produced by humans.  Additionally, a Hierarchical Recurrent Encoder-Decoder (HRED), as a widely-used end-to-end approach for dialogue systems, has been trained as another baseline model. Moreover, we have done several ablation studies to show the importance of each module in our proposed model. 




\begin{table*}[t]
  \caption{Evaluation metrics for language and pricing evaluation of the models. $\uparrow$ indicates higher is better, while $\downarrow$ shows lower is better. 
  \newline* Our full \emph{Price Negotiator} model.
  \newline** Our full \emph{Price Negotiator} when using reinforcement learning.}
  \label{tab:eval}
  \centering
  \resizebox{0.95\linewidth}{!}{
  \begin{tabular}{p{4.5cm}p{0.9cm}p{0.9cm}p{1.2cm}p{1.2cm}p{1.2cm}p{1.7cm}p{1.7cm}p{1.7cm}}
    \toprule
     &  \multicolumn{5}{c}{\textbf{Language Metrics}}&  \multicolumn{3}{c}{\textbf{Pricing Metrics}}\\
    \cmidrule(lr){2-6}\cmidrule(lr){7-9}
    \textbf{Model}&IBLEU$\uparrow$&BLEU$\uparrow$&Sentence Diversity$\uparrow$&Vocabulary Diversity$\uparrow$&Dialogue Length$\downarrow$&Inconsistency in Pricing$\downarrow$&Inconsistency in Offering$\downarrow$&Human Divergence$\downarrow$\\ 
    \midrule
    SL(act)+rule\cite{He2018_craigslist}&   20.21&        2.59&    \textbf{0.498}&   \textbf{0.0467}   &18	        &1\%	  &9\%    &\$383\\
    SL(word)\cite{He2018_craigslist}&        35.76&        3.74&    0.310&   0.0385&            \textbf{7}	        &6\%	  &6\%  &\$375\\
    HRED\cite{Sordoni2015_hred}&                    36.50&          4.56&           0.316&           0.0336&        10&	        6\%	            &17\%       &\$325\\
    \midrule
    OVE+HRNE+LD&                37.32&          4.24&           0.353&           0.0357&         9&	        9\%	            &27\%       &\$293\\
    OVE+HRNE+AP+LD&             39.12&          4.74&           0.375&           0.0358&        11&	 \textbf{0\%}   &\textbf{0\%}       &\$152\\
    OVE+HRNE+AP+PA+LD*&          41.68&  \textbf{4.85}&          0.442&           0.0430&         9&  \textbf{0\%}   &\textbf{0\%}       &\$132\\
    OVE+HRNE+AP+PA+LD+RL**&\textbf{42.90}&        4.65&           0.463&           0.0432&         9&  \textbf{0\%}   &\textbf{0\%}  &\textbf{\$125}\\
  \bottomrule
\end{tabular}}
\vspace{-4mm}
\end{table*}

\begin{table*}[t]
 \caption{Human study results. Turing test shows the rates at which dialogues generated from each model has been identified as human generated negotiations. Comparison results reveal the ratio of preferring the dialogues generated from each model over those from the other one. The last three columns demonstrate the average scores provided by humans after interactive negotiations with models.}\label{tab:human_eval}
 \vspace{-1mm}
  \centering
  \resizebox{.75\linewidth}{!}{
  \begin{tabular}{p{3.2cm}p{1.6cm}p{1.cm}p{1.cm}p{1.cm}p{1.cm}p{1.cm}p{1.cm}}
  \toprule
  \textbf{Model} &  \textbf{Turing Test}&  \multicolumn{3}{c}{\textbf{Comparison Test}}&  \multicolumn{3}{c}{\textbf{Interactive Test}}\\
   \cmidrule(lr){3-5}\cmidrule(lr){6-8}
    &&Human-likeness&Language&Pricing&Human-likeness&Language&Pricing\\
    \midrule
    SL(word)&   35\%&      37\%&      34\%&        31\%&    2.2&    3.2&    2.6\\
    Price Negotiator+RL&      \textbf{49\%}&     \textbf{63\%}&	    \textbf{66\%}&	    \textbf{69\%}&    \textbf{3.8}&    \textbf{4.0}&    \textbf{4.3}\\
  \bottomrule
\end{tabular}
}
\vspace{-5mm}
\end{table*}

\textit{\textbf{Language Evaluation Results.}}
Table~\ref{tab:eval} demonstrates the fact that price elimination from the language vocabulary improves the language quality. Especially, compared with SL(word) and HRED which are not based on human-crafted rules, dialogues generated from both \emph{price negotiator} models enjoy remarkably more language diversity both in word level and sentence level, as the ratio of repetitive sentences, which has been encountered as a common problem in dialogue generation, has increased significantly in both variations of the proposed framework. Additionally, the dialogue and utterance length of the dialogues generated from these models is large enough to show the richness of the generated dialogues. Although it can be inferred from the results that SL(act)+rule is generating linguistically better dialogues as the sentence and vocabulary diversity of this model is larger than the proposed model, it should be mentioned that this diversity is due to heuristic rules that select templates from the dataset that are different from the previously selected ones.

Furthermore, a brief look at the IBELU scores demonstrates the superior performance of \emph{price negotiator} model in comparison to all other ones. It means that this model acts most similarly to humans in different situations. Interestingly, a noticeable improvement in IBLEU score by applying reinforcement learning illuminates that RL pushes the agent to take more human-like decisions in different situations to persuade the opponent and receive the best reward.

Last but not least, the language assessment metrics in Table~\ref{tab:human_eval}--including the Turing test and both human-likeness and language rates for both comparison and interactive tests--indicate that our proposed model has been accepted as a considerably more fluent and human-like negotiator by human participants.

\begin{table}[t]
 \caption{The table shows the average distance of agreed prices by our \textbf{price negotiator} models from their understanding about the value of the item (from OVE). The smaller value indicates the model's ability to insist on achieving its initially estimated value.}\label{tab:agreed_price_eval}
  \vspace{-2mm}

  \centering
  \resizebox{0.9\linewidth}{!}{
  \begin{tabular}{p{1.7cm}p{2.5cm}p{3.2cm}}
  \toprule
    \textbf{Category}&\textbf{Price Negotiator}&\textbf{Price Negotiator+RL}\\
    \midrule
    Bike            &   \$43    &      \textbf{\$40}      \\
    Car             &   \$712    &     \textbf{\$607}       \\
    Electronics     &   \$8    &       \textbf{\$7}     \\
    Furniture       &   \$23    &      \textbf{\$22}      \\
    Housing         &   \$135    &     \textbf{\$117}       \\
    Phone           &   \$20    &      \textbf{\$19}      \\
    \midrule
    Overall           &   \$151    &   \textbf{\$131}         \\
    
  \bottomrule
\end{tabular}
}
\vspace{-7mm}

\end{table}

\textit{\textbf{Pricing Evaluation Results.}}
Table~\ref{tab:eval} demonstrates results of calculating the pricing metrics. It is noticeable that both versions of the \emph{price negotiator} models learn to propose consistent prices while maintaining the language quality. Besides, these models never make a mistake in offering prices which are in conflict with the prices discussed and agreed upon during the dialogue. 
 
More importantly, the proposed \emph{price negotiator} model understands the suitable agreement price for an item precisely. Table~\ref{tab:eval} shows remarkable decrease in agreed price divergence (the difference between the prices agreed by the model with those agreed by humans) resulted from \emph{price negotiator} in comparison to those from other models. In other words, the proposed model can learn the value of items by an online value estimation and reach agreement on prices very close to those agreed by humans by taking human-like actions both in generating utterances and in proposing prices.

Finally, human pricing assessment results in Table~\ref{tab:human_eval} show our proposed \emph{price negotiator} model consistently performs well in comparison to its counterpart.



\textit{\textbf{Results from RL.}}
In Table \ref{tab:eval} and \ref{tab:agreed_price_eval} we observe that using RL generally improves the performance. In particular, in Table \ref{tab:agreed_price_eval} when RL is used, our approach learns to insist on the prior value estimated by OVE more effectively and utilise the language better to achieve its goal (i.e. buying/selling with minimal compromise to that estimated by OVE). 

\textit{\textbf{Ablation Study Results.}}
Table \ref{tab:eval} summarises the effect of each module independently. Simply adding \emph{online value estimator} to the HRED model (OVE+HRNE+LD) results in a slight improvement in both IBLEU and human divergence metrics implying the agent's ability to better imitate human behaviour. Once the \emph{action predictor (AP)} is added we observe significant improvement in negotiation, both linguistically and in agreement prices. 
It is worth noting that action predictor controls the decisions of the agent and leads to reaching agreements that are remarkably closer to those by humans. Furthermore, having \emph{price adjuster (PA)} and putting all of the modules together, not only the language quality is enhanced, but crucially the gap between machine and human agreed prices decreased considerably. This is due to the agent's capability to use a specialised module that handles the prices. Finally, applying \emph{reinforcement learning (RL)} helps the agent to make considerably better performance linguistically and price-wise. It should be stated that since the \emph{language decoder (LD)} is an obligatory component and has the same architecture as other baselines, the ablation study over this component is not been considered.

\vspace{-2mm}
\section{Conclusion and Future Works}

In this paper, we proposed a visual goal-oriented dialogue model for the seller-buyer negotiation. Our model, \emph{Price Negotiator} is a modular framework for negotiation that utilises insights from human's behaviour for disentangling various parts. In particular, we are the first model to incorporate a matching network for evaluating the underlying value of an item by consulting online stores in negotiation.
Experiments on CraigslistBargain dataset show the superior performance of the proposed model both linguistically  and in reaching a human-like agreement price in various scenarios.

For future we consider improving the current approach by: (a) adding external knowledge about the cost and the availability of side-offers, like free delivery; and (b) applying pre-trained language models, such as BERT \cite{Devlin018_bert}, that may improve the understanding and generation performance.











\bibliographystyle{unsrt}
\bibliography{bibliography}


\newpage
\begin{figure*}
  \centering
  \includegraphics[width=\textwidth]{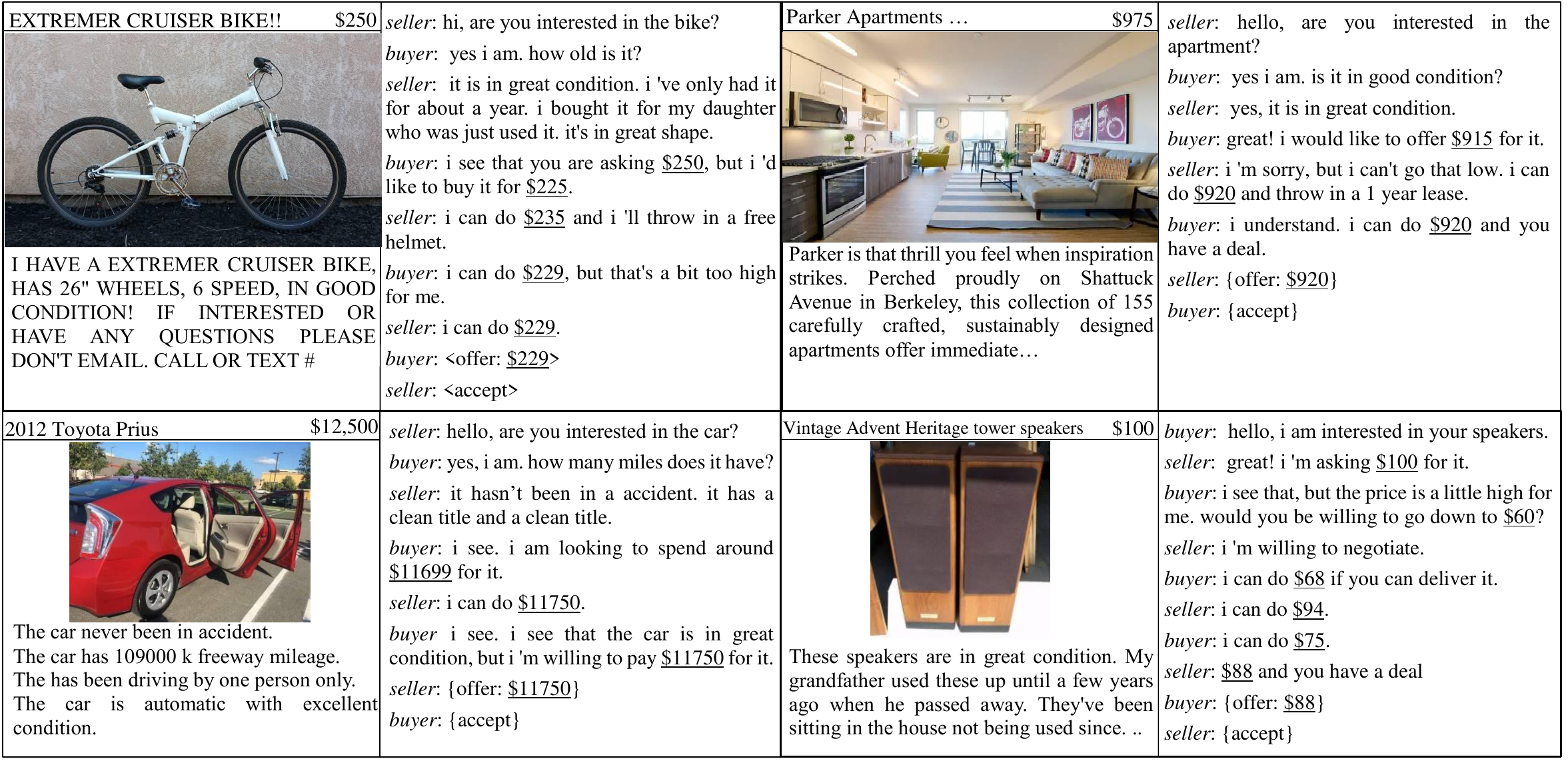}
  \vspace{-5mm}
  \caption{More samples from our \emph{price negotiator} model.}
  \label{fig:mode_samples}
\end{figure*}

\section{Supplementary Materials}

\subsection{Estimated Value Usage}
To have a better understanding about the importance of the estimated value (from OVE) in the negotiation, we assess the agent's performance according to the changes in the estimated value. To that end, during the test time of the \emph{Visual Negotiator+RL} agent, we make the negotiation more challenging by either increasing the estimated value for the seller (by 20\%) or decreasing it for the buyer (again by 20\%). 

Interestingly, this modification results in a considerable increase in the distance between the agreed prices and the estimated value of the item (from \$131 to around \$145). Additionally, we observe a 2\% decrease in the agreement ratio of negotiations. These results reveal the impact of the value estimation on the agent and show that the agent competes with its counterpart to reach the most beneficial deal according to this estimated value, even if this competition ends up with a disagreement. 

\subsection{Human Evaluation Web Interface}
\begin{figure*}
  \centering
  \includegraphics[width=\textwidth]{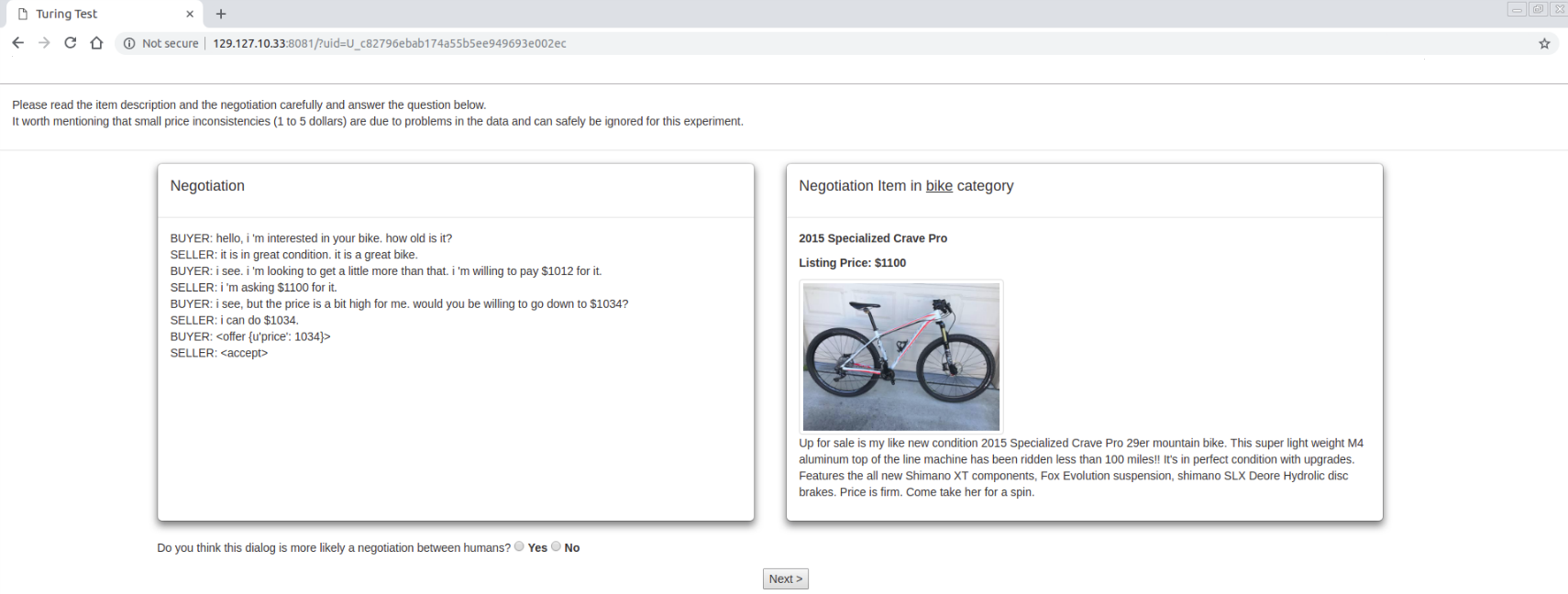}
  \vspace{-5mm}
  \caption{\small{A picture of our Turing test web page where humans can read a negotiation over an item and choose whether the dialogue has been produced by humans or not.}}
  \label{fig:TuringTest}
\end{figure*}

\begin{figure*}
  \centering
  \includegraphics[width=\textwidth]{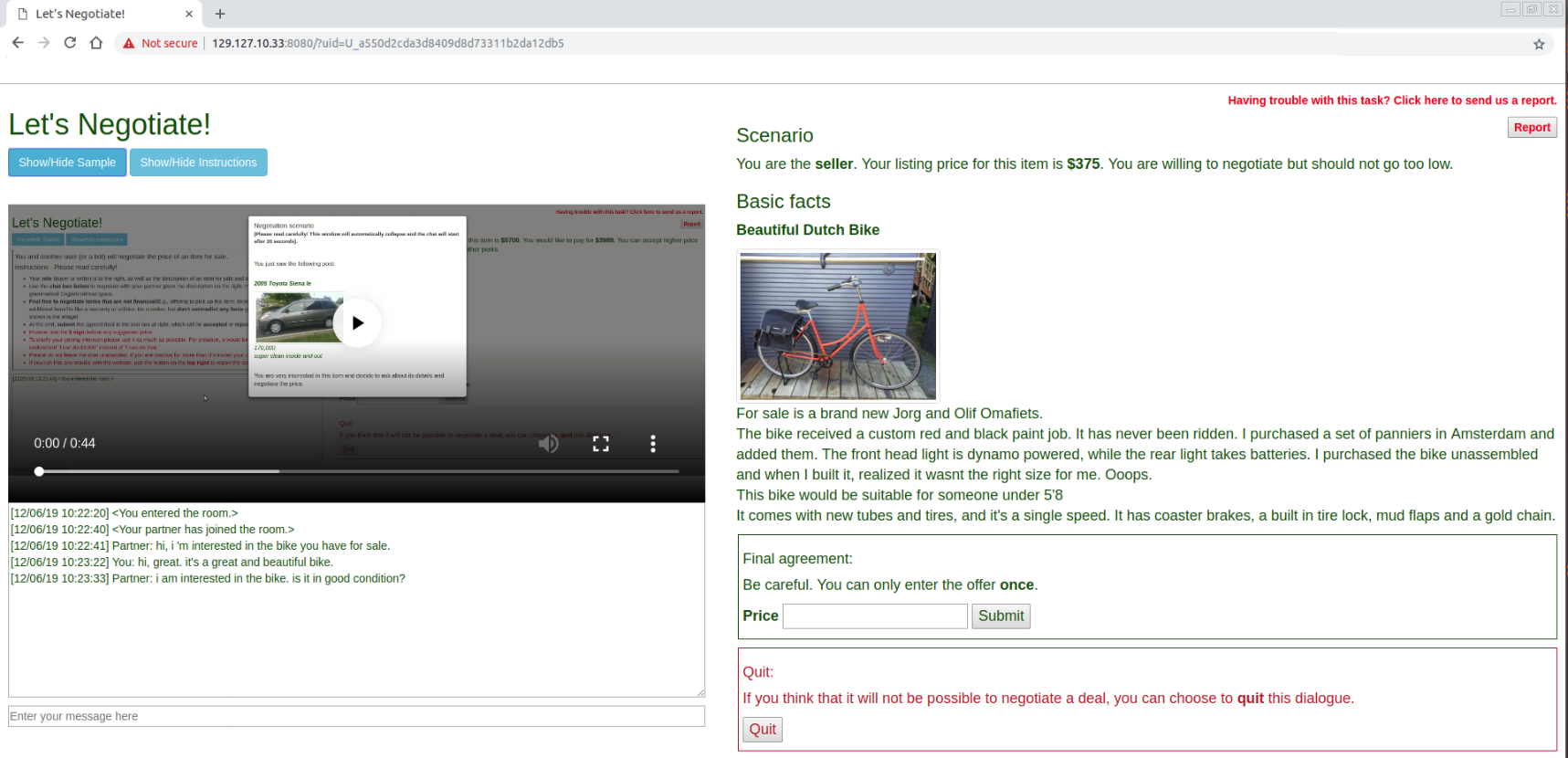}
  \vspace{-5mm}
  \caption{\small{A snapshot of our interactive human evaluation. After negotiating over an item with a chat-bot, humans are asked to rate its performance. The website is developed by borrowing code from CoCoA framework (\url{https://github.com/stanfordnlp/cocoa).}}}
  \label{fig:LiveChat}
\end{figure*}

Snapshots of the web interfaces of different human evaluations are shown in Figure~\ref{fig:ComparisonTest} (\emph{Comparative Test}), Figure~\ref{fig:TuringTest} (\emph{Turing Test}), and Figure~\ref{fig:LiveChat} (\emph{Interactive Test}). Additionally, a sample video of a live chat between our Price Negotiator and a human is provided through this link: \url{https://drive.google.com/file/d/1tiKCo2ven0clScLtUsNWOwiDahYy4iXf/view?usp=sharing}

\begin{figure*}
  \centering
  \includegraphics[width=\textwidth]{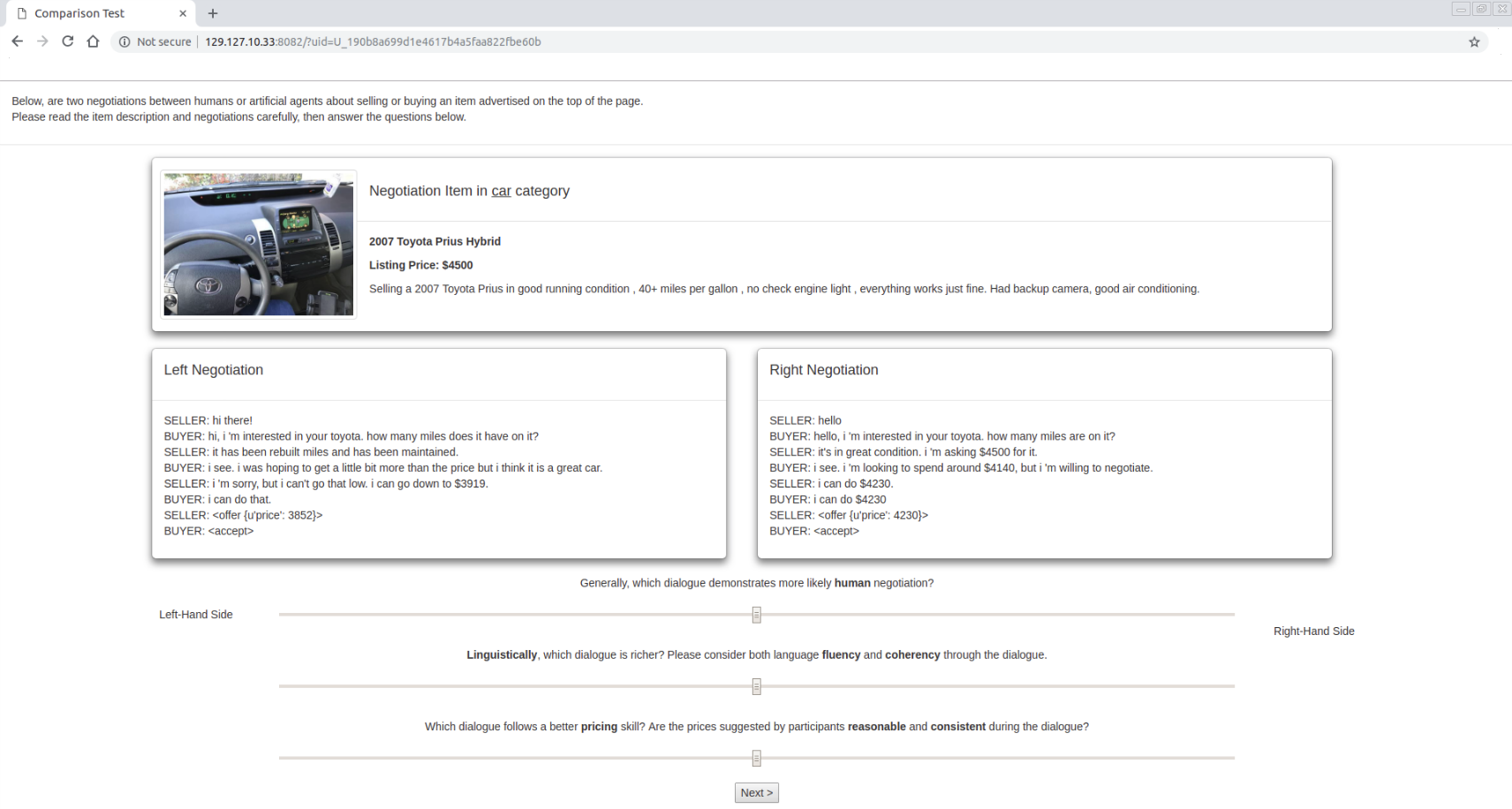}
  \vspace{-5mm}
  \caption{\small{A snapshot of our online human evaluation in which the participant is given two negotiations generated from different models over the same item and is asked to choose the better one in terms of human likeness, language fluency and pricing quality.}}
  \label{fig:ComparisonTest}
\end{figure*}

\end{document}